\documentclass[10pt,twocolumn,letterpaper]{article}

\usepackage[margin=0.75in,top=1in]{geometry}
\usepackage{placeins}

\usepackage{algorithm}
\usepackage{algorithmic}
\usepackage{multicol, multirow}
\usepackage{cuted}
\usepackage{capt-of}
\usepackage{times}
\usepackage{graphicx}
\usepackage{amsmath,amssymb,amsfonts}
\usepackage{booktabs}
\usepackage[dvipsnames]{xcolor}
\usepackage{subcaption}
\usepackage{bm}

\definecolor{linkblue}{rgb}{0.21,0.49,0.74}
\usepackage[pagebackref,breaklinks,colorlinks,allcolors=linkblue]{hyperref}

\title{MatRes: Zero-Shot Test-Time Model Adaptation\\for Simultaneous Matching and Restoration}

\author{
Kanggeon Lee$^{1}$, Soochahn Lee$^{2}$, Kyoung Mu Lee$^{1}$ \\
$^{1}$ASRI, Dept. of ECE, Seoul National University, Korea \\
$^{2}$Dept. of Electronics Engineering, Kookmin University, Korea \\
\footnotesize\texttt{dlrkdrjs97@snu.ac.kr, sclee@kookmin.ac.kr, kyoungmu@snu.ac.kr}
}

\date{}

\begin{document}

\maketitle
\begin{strip}
    \centering
    \includegraphics[width=\textwidth]{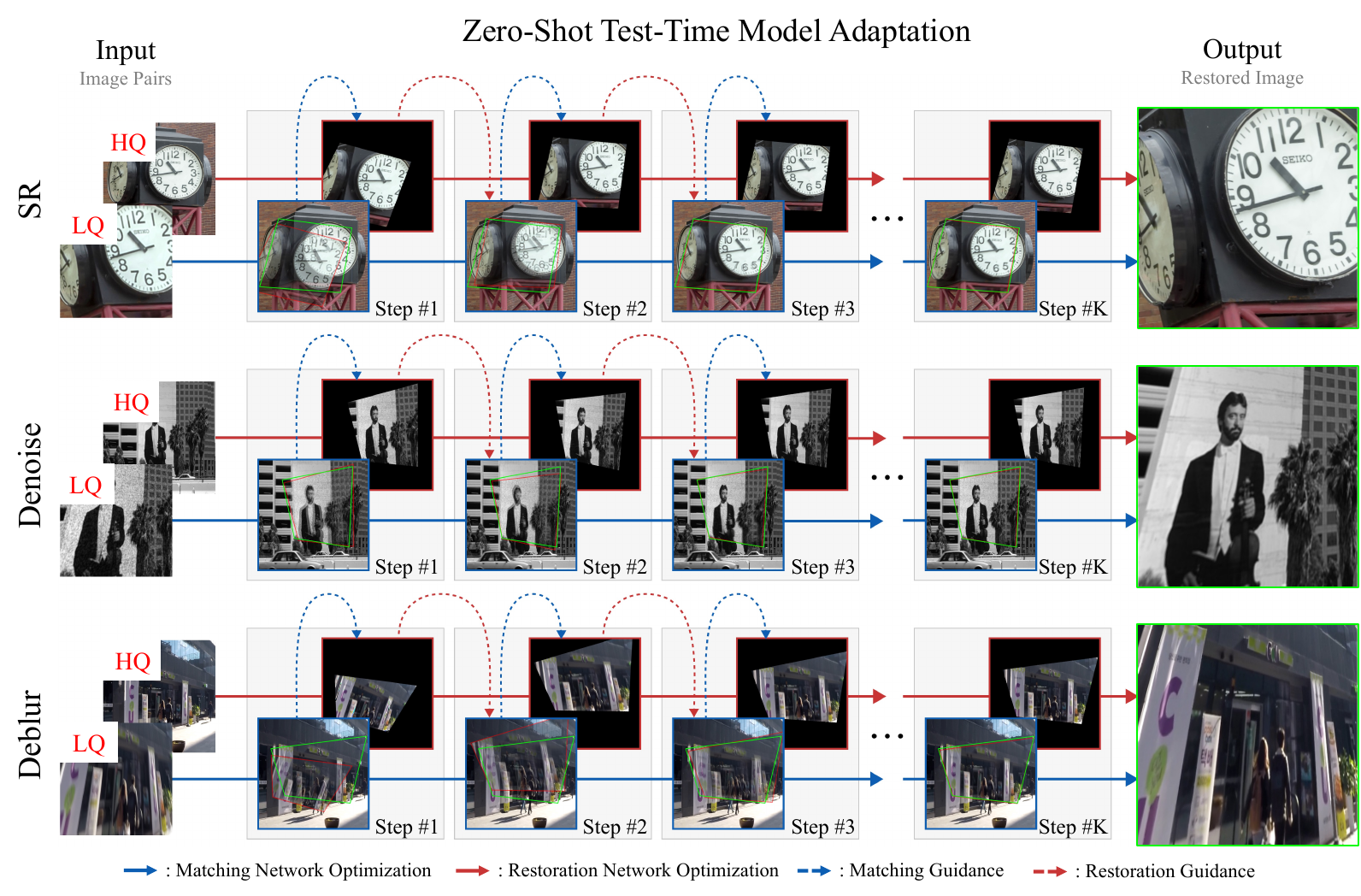}
    \vspace{2mm}
    \captionof{figure}{\textbf{Zero-shot Test-Time Adaptation.} \textsc{MatRes} enhances degraded and viewpoint-shifted image pairs by leveraging the mutual guidance of a matching and a restoration network to produce a restored and aligned output image.}
    \label{fig:FIG_TEASER}
\end{strip}

\begin{abstract}
Real-world image pairs often exhibit both severe degradations and large viewpoint changes, making image restoration and geometric matching mutually interfering tasks when treated independently. 
In this work, we propose \textsc{MatRes}, a zero-shot test-time adaptation framework that jointly improves restoration quality and correspondence estimation using only a single low-quality and high-quality image pair. 
By enforcing conditional similarity at corresponding locations, \textsc{MatRes} updates only lightweight modules while keeping all pretrained components frozen, requiring no offline training or additional supervision. 
Extensive experiments across diverse combinations show that \textsc{MatRes} yields significant gains in both restoration and geometric alignment compared to using either restoration or matching models alone. 
\textsc{MatRes} offers a practical and widely applicable solution for real-world scenarios where users commonly capture multiple images of a scene with varying viewpoints and quality, effectively addressing the often-overlooked mutual interference between matching and restoration.
\end{abstract}    
\section{Introduction}
\label{sec:intro}
When photographing a scene, it is common to capture multiple images from different viewpoints and poses. 
In practice, however, some of these images often suffer from degradations such as blur, noise, or low light. 
Under such real-world conditions, the two fundamental tasks, restoring image content and establishing accurate correspondences, tend to hinder each other when performed independently.

To fully exploit the available images, restoration and matching should ideally be performed jointly so that each task can benefit from the other. 
However, despite extensive research on image matching~\cite{haskins2020deep,xu2024local,zhang2025deep} and image restoration~\cite{fan2019brief,wang2020deep,zhang2022deep,su2022survey,elad2023image,zhai2023comprehensive,jiang2025eficient}, including recent all-in-one restoration approaches typically designed for single images~\cite{jiang2025survey}, few methods address their joint optimization.


Some methods have proposed to perform image alignment and restoration jointly or iteratively, but for grouped images with very small baselines~\cite{zhang2014multi,bhat2021deep,lecouat2021lucas,guo2022differentiable} or videos~\cite{sellent2016stereo,kim2018spatio,pan2019joint,Tian_2020_CVPR}.
To the best of our knowledge, there has yet to be sufficient research on this approach for image pairs with wide baselines, as those in image matching benchmarks~\cite{balntas2017hpatches,lin2015microsoftcoco,zhao2021deep,li2018megadepthlearningsingleviewdepth,dai2017scannet}.
This lack of research is most likely attributed to the dilemma of data acquisition.
While the complexity of the task warrants a deep learning approach, it is nearly impossible to acquire the training data in real-world situations suitable to achieve generalization and robust performance, including ground-truth (GT) clean images, GT transformation fields, and known degradation kernels.

In this paper, we propose a novel framework to address the issue of simultaneous matching and restoration, which we call \textsc{MatRes}.
The key to our framework is based on the realistic assumption that for the given image pair, one is the more corrupted, low-quality (LQ) image while the other is the relatively high-quality (HQ) image that can be used as reference.
Then, pretrained models for matching and restoration can be adapted to the given images by enforcing conditional similarity between the LQ and HQ images at corresponding points.
Our method eliminates the need for offline training, performing adaptation through an iterative optimization that functions directly at inference time.

Our method comprises three sub-networks: a pretrained matching network, a pretrained restoration network, and a lightweight LoRA-based projection network~\cite{hu2022lora}. 
Given an input LQ--HQ image pair, the matching network first extracts generative priors and estimates the geometric relationship between the two images. 
These priors are then projected into the feature space of the restoration network by the LoRA projection network, providing alignment-aware conditioning for restoration. 
The loss function comprises two terms; 
the sparsity of the similarity matrix of the feature maps at coinciding local neighborhoods, and the pixel differences between the restored images in overlapping positions. 
These loss terms are used to iteratively train the LoRA parameters.
Finally, the restoration network generates the restored image, while the output of the matching network generates the image correspondences. 

We present extensive experimental results with different combinations of various matching and restoration networks.
We demonstrate that this mutual guidance between matching and restoration network yields substantial improvements in restoration quality compared to applying the restoration network to the LQ image alone without any geometric guidance.
We also demonstrate that the accuracy of point correspondences of our framework significantly outperforms the results of using the matching network alone.

Overall, our work provides the following contributions:
\begin{itemize}
    \item The proposed method is practical and widely applicable for real-world scenarios, as it is very common for the user to take a group of two or more images of the scene with varying viewpoints and quality, and addresses the issue of mutual deterrence of matching and restoration, which has mostly been overlooked by previous works.
    \item Our method does not require any additional training data other than the given reference image in the input image pair, and does not require any offline training.
    \item Our method jointly improves both restoration quality and matching accuracy compared to existing methods that separately perform each task, for a pair of images. 
\end{itemize}

\section{Related works}
\label{sec:rel}

\textbf{Joint matching and restoration methods}
The work of Zhang et al.~\cite{zhang2014multi} is a notable attempt at addressing the issue of simultaneous restoration and matching, prior to deep learning. 
A mathematical model is formulated on the assumed relation between image transforms and blur kernels, both assumed to be based on the motion between the images, and iteratively optimized to recover the restored image and the motion parameters.
While the method provides a sophisticated approach, it is limited to images taken at very close intervals. 
More recent works employing deep learning have also leaned into this assumption, operating on burst mode photos~\cite{,lecouat2021lucas,guo2022differentiable} or video~\cite{sellent2016stereo,kim2018spatio,pan2019joint,Tian_2020_CVPR}.
Our method differs with these works, as we assume image pairs where the degradations are uncorrelated with the geometric transforms between images.
Moreover, the transforms can be formulated to be non-rigid, depending on the component matching model.

\textbf{Restoration and matching methods}
\textsc{MatRes} is compatible with any pretrained matching model where a gridwise feature map is computed, from which the geometric transform, most commonly in a deformation field format, in generated.
This includes recent deep learning–based correspondence methods such as LoftUp~\cite{huang2025loftuplearningcoordinatebasedfeature}, ART~\cite{Lee_2025_ICCV}, 
DIFT~\cite{Tang2023EmergentCorrespondence}, RoMa~\cite{edstedt2024roma}, and Mast3R~\cite{mast3r_eccv24}. 
Similarly, any restoration model equipped with an encoder that produces gridwise feature representations and a decoder that reconstructs the restored image is compatible with \textsc{MatRes}. 
This covers virtually all modern restoration architectures, including general-purpose models such as Restomer~\cite{Zamir2021Restormer} and SwinIR~\cite{liang2021swinir}, as well as task-specific methods for super-resolution, denoising, and deblurring, such as EDSR~\cite{Lim_2017_CVPR_Workshops}, N2N~\cite{lehtinen2018noise2noiselearningimagerestoration}, N2V~\cite{krull2019noise2void}, DeblurGAN~\cite{DeblurGAN}, and SRN~\cite{tao2018srndeblur}.

\begin{figure}[!htbp]
    \centering
    \includegraphics[width=1.0\columnwidth]{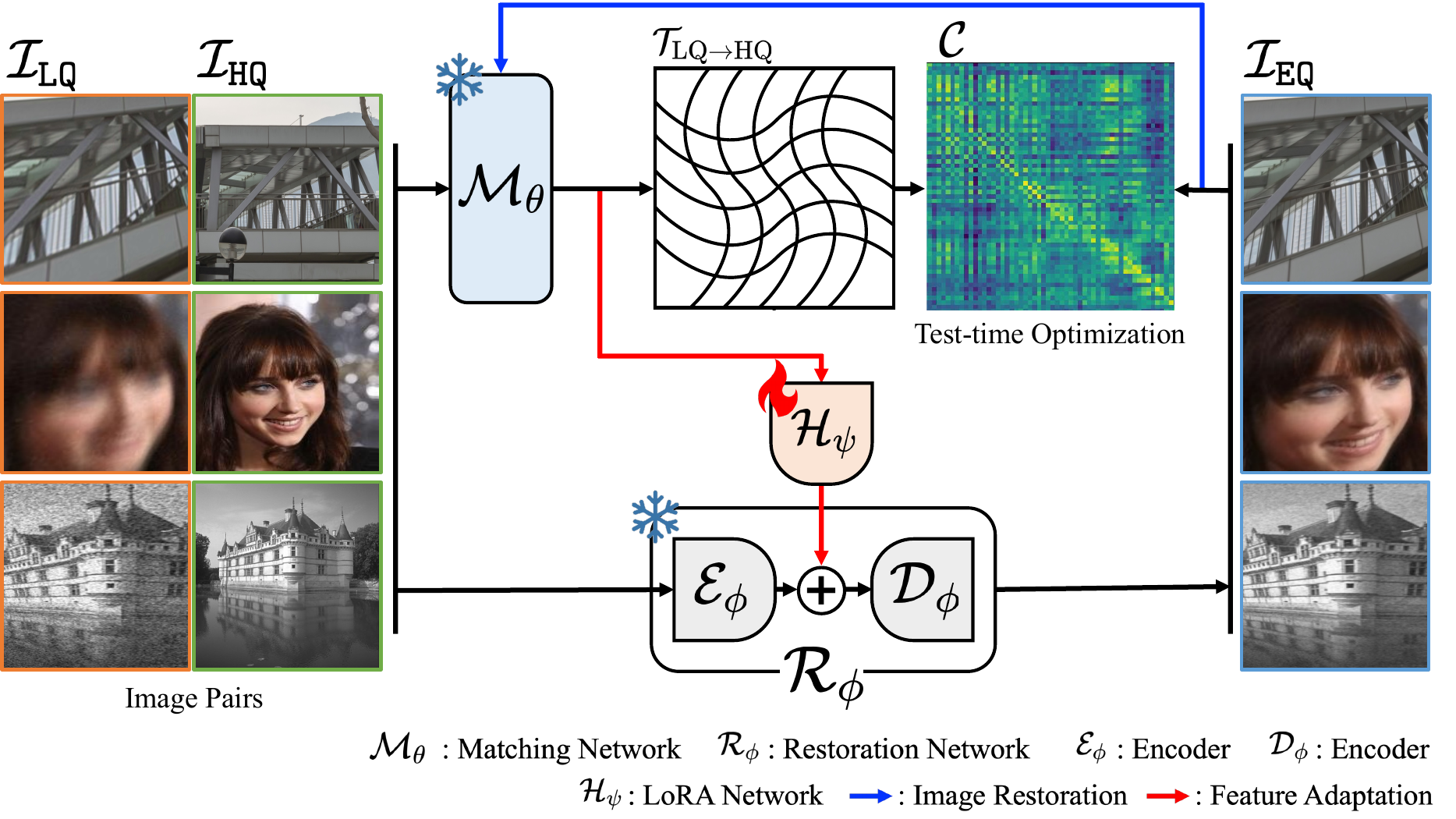}
    \caption{
    \textbf{Method Overview.} Given an input pair $(\mathcal{I}_\texttt{LQ}, \mathcal{I}_\texttt{HQ})$, the pretrained generative prior model $\mathcal{M}_\theta$ and the pretrained restoration network $\mathcal{R}_\phi$ remain frozen during adaptation (gradient flows through both models but their parameters are not updated).
    The zero-initialized adapter $\mathcal{H}_\psi$ is the only trainable module, refined iteratively using losses computed on the $\mathcal{I}_\texttt{EQ}$ and $\mathcal{I}_\texttt{HQ}$ image pairs. At each adaptation step, $\mathcal{M}_\theta$ extracts generative priors and estimates the transformation $\mathcal{T}_{\texttt{LQ}\rightarrow\texttt{HQ}}$, the $\mathcal{I}_\texttt{LQ}$ image are warped and injected into $\mathcal{R}_\phi$, and the adapter $\mathcal{H}_\psi$ learns to align and modulate features for improved restoration.
    } 
    \label{fig:FIG_NETWORK}
\end{figure}

\begin{algorithm}[t!]
\caption{Test-time Zero-shot Adaptation}
\label{alg:tta}
\begin{algorithmic}[1]
\REQUIRE Input pair $(\mathcal{I}_\texttt{LQ}, \mathcal{I}_\texttt{HQ})$, pretrained $\mathcal{M}_\theta$, pretrained $\mathcal{R}_\phi$, zero-initialized adapter $\mathcal{H}_\psi$
\ENSURE Updated adapter parameters $\psi$

\FOR{$t = 1$ to $\textsc{T}_{\text{adapt}}$}
    \STATE \textbf{Extract generative priors from $\mathcal{M}_\theta$:} $z_\texttt{LQ}, z_\texttt{HQ}$
    \STATE \textbf{Estimate transform from priors:} $\mathcal{T}_{\texttt{LQ}\rightarrow\texttt{HQ}} = \mathcal{M}_\theta(z_\texttt{LQ}, z_\texttt{HQ})$
    \STATE Warp $\mathcal{I}_\texttt{LQ}$ and $z_\texttt{LQ}$ using $\mathcal{T}_{\texttt{LQ}\rightarrow\texttt{HQ}}$
    \STATE Adapt features: $\tilde{z}^{\texttt{WARPED}}_\texttt{LQ} = \mathcal{H}_\psi(z^{\texttt{WARPED}}_\texttt{LQ})$, \ \ $\tilde{z}_\texttt{HQ} = \mathcal{H}_\psi(z_\texttt{HQ})$
    \STATE Restore: $\mathcal{I}_\texttt{EQ} = \mathcal{R}_\phi(\mathcal{I}^{\texttt{WARPED}}_\texttt{LQ}, \mathcal{I}_\texttt{HQ} \mid \tilde{z}^{\texttt{WARPED}}_\texttt{LQ}, \tilde{z}_\texttt{HQ})$
    \STATE Compute cost volume: $\mathcal{C} = {(\hat z^{\texttt{WARPED}}_\texttt{EQ})}^\top \hat z_\texttt{HQ}$
    \STATE Compute off-diagonal and pixel-matching losses on the warped region
    \STATE Update adapter parameters $\psi$ via gradient descent while keeping $\mathcal{M}_\theta$ and $\mathcal{R}_\phi$ frozen
\ENDFOR
\RETURN $\psi$
\end{algorithmic}
\end{algorithm}

\section{Method}
Fig.~\ref{fig:FIG_NETWORK} provides an overview of our proposed test-time zero-shot adaptation method.
Our proposed zero-shot adaptation algorithm comprises three networks: an off-the-shelf 2D image matching network $\mathcal{M}_\theta$, a restoration model $\mathcal{R}_\phi$ with encoder \(\mathcal{E}_\phi\) and decoder \(\mathcal{D}_\phi\) combination, and a LoRA $\mathcal{H}_\psi$, where the parameters $\theta$ and $\phi$ are fixed but allow gradient flow, and $\psi$ is learnable.
We begin by describing how to extract generative priors from various types of matching models $\mathcal{M}_\theta$, and then explain how to inject these priors into any image restoration model using a LoRA $\mathcal{H}_\psi$, along with the procedure for training $\mathcal{H}_\psi$ at test time.

\subsection{Problem Formulation}
Given a pair of input images $\mathcal{I}_\texttt{LQ}$ and $\mathcal{I}_\texttt{HQ}$, referred to as the LQ and HQ images respectively, the goal is to estimate the geometric transformation $\mathcal{T}_{\texttt{LQ} \rightarrow \texttt{HQ}}$ that aligns $\mathcal{I}_\texttt{LQ}$ to $\mathcal{I}_\texttt{HQ}$, while simultaneously enhancing the quality of $\mathcal{I}_\texttt{LQ}$ to generate the image $\mathcal{I}_\texttt{EQ}$ (EQ denoting enhanced quality).
To this end, matching network $\mathcal{M}_\theta$ and restoration network $\mathcal{R}_\phi$ which produce the predicted transformation  and the enhanced image, respectively, are employed.
However, in practice, $\mathcal{M}_\theta$ often suffers from severe performance degradation when applied to new domains or previously unseen datasets. 
Fine-tuning the model on such novel data typically requires ground-truth correspondences, which are unavailable in test-time scenarios, making zero-shot adaptation particularly challenging.
Similarly, in most real-world scenarios, $\mathcal{R}_\phi$ cannot effectively enhance $\mathcal{I}_\texttt{LQ}$ from a misaligned $\mathcal{I}_\texttt{HQ}$. 
Although a high-quality reference image $\mathcal{I}_\texttt{HQ}$ is available, it remains unusable without precise spatial alignment between the two modalities.

\subsection{Mutual Guidance}
Recent studies~\cite{Tang2023EmergentCorrespondence, Zhang2023TaleOfTwoFeatures, Cheng2024ZeroShotImageFeatureConsensus, SIVARAMAN2025109645} have revealed that large-scale matching models $\mathcal{M}_\phi$ exhibit surprisingly strong emergent correspondence properties which can act as a generative prior. 
In particular, features extracted from such models can enable accurate nearest-neighbor matching even without task-specific supervision. 
These generative features are learned from large and diverse datasets, and thus often generalize well to unseen domains~\cite{SIVARAMAN2025109645}.

Motivated by this, we propose to leverage the informative and domain-generalizable features from the matching model $\mathcal{M}_\theta$ to facilitate the restoration process. 
Specifically, the estimated transformation $\mathcal{T}_{\texttt{LQ} \rightarrow \texttt{HQ}}$ derived from $\mathcal{M}_\theta$ provides spatial guidance from the high-quality reference $\mathcal{I}_\texttt{HQ}$, enabling the restoration network $\mathcal{R}_\phi$ to generate an enhanced image $\mathcal{I}_\texttt{EQ}$ that is more consistent with $\mathcal{I}_\texttt{HQ}$. 
In turn, this refined $\mathcal{I}_\texttt{EQ}$ further improves the alignment process, establishing a mutual reinforcement between matching and restoration, all without requiring ground-truth correspondences.

Building upon this idea, we extract generative features $z_\texttt{LQ}$ and $z_\texttt{HQ}$ from $\mathcal{M}_\theta$, compute geometric transformation $\mathcal{T}_{\texttt{LQ} \rightarrow \texttt{HQ}}$ from extracted generative features, and inject them into the inference pipeline of $\mathcal{R}_\phi$, modifying the transformation prediction as:
\[
\mathcal{I}^{\texttt{WAPRED}}_\texttt{EQ} = \mathcal{R}_\phi(\mathcal{I}^{\texttt{WAPRED}}_\texttt{LQ}, \mathcal{I}_\texttt{HQ} \mid \tilde{z}^{\texttt{WAPRED}}_\texttt{LQ}, \tilde{z}_\texttt{HQ}),
\]
where $\mathcal{I}^{\texttt{WAPRED}}_\texttt{LQ}$ and $z^{\texttt{WAPRED}}_\texttt{LQ}$ denote the warped LQ image and its corresponding warped generative feature obtained by applying the estimated transformation $\mathcal{T}_{\texttt{LQ} \rightarrow \texttt{HQ}}$ predicted from $\mathcal{M}_\theta$, respectively. 
The adapted features $\tilde{z}^{\texttt{WAPRED}}_\texttt{LQ}= \mathcal{H}_\psi(z^{\texttt{WAPRED}}_\texttt{LQ})$ and $\tilde{z}_\texttt{HQ}= \mathcal{H}_\psi(z_\texttt{HQ})$ are produced by a learnable feature projector $\mathcal{H}_\psi$, which aligns the generative priors from the matching model with the internal feature space of the restoration network, effectively bridging the distributional gap between $\mathcal{M}_\theta$ and $\mathcal{R}_\phi$.

Then, we can update $\mathcal{I}_\texttt{LQ}$ to $\mathcal{I}_\texttt{EQ}$ for following iteration.


\subsection{Generative Prior Extraction}
Here, we describe how to extract generative priors from different types of matching networks.
\paragraph{Diffusion Models}
are generative frameworks that learn to reverse a gradual noising process. 
In the context of latent diffusion~\cite{Song2021DDIM, Rombach2022LDM}, the forward process corrupts a clean latent feature $z_0$ from input image $\mathcal{I}$ using the encoder of latent diffusion model into a sequence $\{z_t\}_{t=1}^T$ through Gaussian perturbations, governed by:
\begin{equation}
    q(z_t \mid z_{t-1}) = \mathcal{N}(\sqrt{1 - \beta_t} \cdot z_{t-1}, \beta_t \mathbf{I}),
\end{equation}
where $\beta_t$ denotes the noise schedule. This can be compactly written as a one-step transformation from $z_0$:
\begin{equation}
    z_t = \sqrt{\bar{\alpha}_t} z_0 + \sqrt{1 - \bar{\alpha}_t} \cdot \epsilon_t, \quad \epsilon_t \sim \mathcal{N}(0, \mathbf{I}),
\end{equation}
with $\bar{\alpha}_t = \prod_{s=1}^t (1 - \beta_s)$.

The reverse denoising path is learned using a U-Net denoiser $\epsilon_\theta(z_t, t)$, which estimates the added noise. Each step reconstructs $z_{t-1}$ as:
\begin{equation}
    z_{t-1} = \frac{1}{\sqrt{\alpha_t}} \left( z_t - \frac{1 - \alpha_t}{\sqrt{1 - \bar{\alpha}_t}} \epsilon_\theta(z_t, t) \right) + \sigma_t \cdot \mathbf{z}, \quad \mathbf{z} \sim \mathcal{N}(0, \mathbf{I}).
\end{equation}

In this paper, we denote the denoised latent diffusion feature $z_{t-1}$ as $z = \mathcal{M}_\theta(\mathcal{I})$, where $\mathcal{I}$ is the input image.
Following prior works~\cite{Tang2023EmergentCorrespondence, Zhang2023TaleOfTwoFeatures, Cheng2024ZeroShotImageFeatureConsensus, SIVARAMAN2025109645}, the denoised latent feature $z$ can be utilized for correspondence search, as it captures high-level semantic information while preserving local structural details. 
Even without explicit supervision, these features naturally exhibit correspondence-friendly behavior, allowing reliable estimation through nearest-neighbor matching in the feature space.

\paragraph{Auto-Regressive Model}
generates data by sequentially predicting visual ques conditioned on previously generated ones, effectively capturing complex spatial dependencies in a causal manner~\cite{tian2024visualautoregressivemodelingscalable}. 
Given an input image $\mathcal{I}$, an encoder in the model produces a hierarchy of multi-scale features $\{f^l\}_{l=1}^L$, where each level $l$ encodes structural and semantic information at different spatial resolutions.
To obtain a compact generative prior representation, we aggregate these multi-scale features by spatially averaging and concatenating them across levels, forming the latent descriptor:
\begin{equation}
    z = \frac{1}{L} \sum_{l=1}^{L} \text{Pool}(f^l),
\end{equation}
where $\text{Pool}(\cdot)$ denotes global average pooling. 
The resulting $z$ serves as a holistic generative feature embedding that summarizes the AR model’s multi-scale representations, providing a semantically rich prior for downstream matching or restoration tasks.

\paragraph{Vision Transformer}
maps an image $x$ into a set of patch-level tokens $z = \{z_i\}_{i=1}^N$, where each $z_i \in \mathbb{R}^d$ encodes information about the corresponding local region of the image~\cite{dosovitskiy2021imageworth16x16words}:
\begin{equation}
    z = \mathcal{M}_\theta(x), \quad z \in \mathbb{R}^{N \times d},
\end{equation}
where $N$ is the number of non-overlapping patches and $d$ is the embedding dimension.

\subsection{Feature Adaptation}
To enable zero-shot test-time adaptation, we introduce a lightweight feature adapter network $\mathcal{H}_\psi$, implemented as a simple convolutional module. 
This adapter refines the generative features produced by the matching model $\mathcal{M}_\theta$ before forwarding them to the restoration network $\mathcal{R}_\phi$.

The adapted features $\tilde{z}^{\texttt{WAPRED}}_\texttt{LQ} = \mathcal{H}_\psi(z^{\texttt{WAPRED}}_\texttt{LQ})$ and $\tilde{z}_\texttt{HQ}= \mathcal{H}_\psi(z_\texttt{HQ})$ are injected into the intermediate feature maps of $\mathcal{R}_\phi$ through additive modulation to create $\mathcal{I}^{\texttt{WAPRED}}_\texttt{EQ}$, allowing the restoration network to effectively exploit the spatial priors encoded in the matching model as:
\[
\mathcal{I}^{\texttt{WAPRED}}_\texttt{EQ} = \mathcal{D}_\phi(\mathcal{E}_\phi(\mathcal{I}^{\texttt{WAPRED}}_\texttt{LQ}) + \tilde{z}^{\texttt{WAPRED}}_\texttt{LQ}, \mathcal{E}_\phi(\mathcal{I}_\texttt{HQ}) + \tilde{z}_\texttt{HQ}).
\]

A key design choice is the zero-initialization of $\mathcal{H}_\psi$, ensuring that the adapter initially outputs zeros and thus does not alter the behavior of the pre-trained $\mathcal{R}_\phi$. 
During test-time optimization, $\mathcal{H}_\psi$ gradually learns to inject informative corrections into the restoration process, progressively tuning $\mathcal{R}_\phi$ to the characteristics of each image pair without modifying the backbone parameters.

This plug-and-play adaptation mechanism enables zero-shot operation by updating only the adapter network $\mathcal{H}_\psi$, while keeping both $\mathcal{M}_\theta$ and $\mathcal{R}_\phi$ frozen. 
As a result, the system can flexibly generalize to unseen domains or modalities, leveraging the priors from $\mathcal{M}_\theta$ for adaptive restoration at test time.

\subsection{Test-time Optimization}
\paragraph{Cost volume.}
At test time, we adapt the feature projector \(\mathcal{H}_\psi\) by minimizing a feature alignment loss between the warped source image and the target image. To this end, we utilize the matching model \(\mathcal{M}_\theta\) to extract features from both images and compute a cost volume that reflects their spatial correspondence.

Given a EQ image $\mathcal{I}_\texttt{EQ}$ and a HQ image $\mathcal{I}_\texttt{HQ}$, we first extract generative features  $z^{\texttt{WARPED}}_\texttt{EQ}$ and $z_\texttt{HQ}$ from $\mathcal{M}_\theta$.
Using these features, we compute a cost volume via batch-wise matrix multiplication between the normalized features:
\[
\mathcal{C} = {(\hat{z}_{\texttt{EQ}}^{\texttt{WARPED}})}^\top \hat{z}_{\texttt{HQ}} \in \mathbb{R}^{HW \times HW},
\]
where $\hat z^{\texttt{WARPED}}_\texttt{EQ} = \textsc{Normalize}(\textsc{Flatten}(\hat z^{\texttt{WARPED}}_\texttt{EQ}))$ and $\hat z_\texttt{HQ} = \textsc{Normalize}(\textsc{Flatten}(z_\texttt{HQ}))$. 
Here, the \(\textsc{Flatten}(\cdot)\) reshapes each feature tensor into a matrix of shape $C \times HW$, 
and \(\textsc{Normalize}(\cdot)\) is the channel-wise normalization.
To stabilize the optimization and ensure consistent scaling, we apply min-max normalization:
\[
\mathcal{C}_{\text{norm}} = \frac{\mathcal{C} - \min(\mathcal{C})}{\max(\mathcal{C}) - \min(\mathcal{C}) + \epsilon},
\]
where \(\epsilon\) is a small constant to prevent division by zero.

\paragraph{Loss function for optimization.}
The end-to-end training loss \(\mathcal{L}\) is defined as:
\begin{equation}\label{eq_full}
\mathcal{L} = \mathcal{L}_{\text{D}} + \lambda_{\text{P}}\mathcal{L}_{\text{P}},
\end{equation}
where \(\mathcal{L}_\text{D}\) and \(\mathcal{L}_\text{P}\) are the off-diagonal loss, and pixel-matching loss, respectively. 
\(\lambda_{\text{P}}\) controls the relative importance of the pixel-matching loss.

We formulate an off-diagonal loss~\cite{Cheng2024ZeroShotImageFeatureConsensus} that penalizes high off-diagonal responses in the cost volume, thereby encouraging sharper, more diagonal-dominant similarity:
\[
\mathcal{L}_\text{D} = \left\| \mathcal{C}_{\text{norm}} - \textsc{Diag}(\mathcal{C}_{\text{norm}}) \right\|_1^1,
\]
where \(\textsc{Diag}(\cdot)\) returns a matrix containing only the diagonal elements of its input, with all off-diagonal entries set to zero.

For pixel-level consistency, we define the pixel-matching loss as the mean squared error between the enhanced image $\mathcal{I}_\texttt{EQ}$ and the high-quality reference $\mathcal{I}_\texttt{HQ}$:
\[
\mathcal{L}_\text{P} = \left\| \mathcal{I}_\texttt{EQ} - \mathcal{I}_\texttt{HQ} \right\|_2^2.
\]

Note that both the off-diagonal loss and the pixel-matching loss are computed only within the warped region, where valid correspondences between $\mathcal{I}_\texttt{EQ}$ and $\mathcal{I}_\texttt{HQ}$ are defined.


Finally, the adapter parameters \(\psi\) are updated via gradient descent:
\[
\psi \leftarrow \psi - \eta \nabla_\psi \mathcal{L}.
\]
This adaptation is repeated for a fixed number of iterations \(\textsc{T}_{\text{adapt}}\), as outlined in Algorithm~\ref{alg:tta}.
\begin{figure*}[!htbp]
    \centering
    \includegraphics[width=1.0\textwidth]{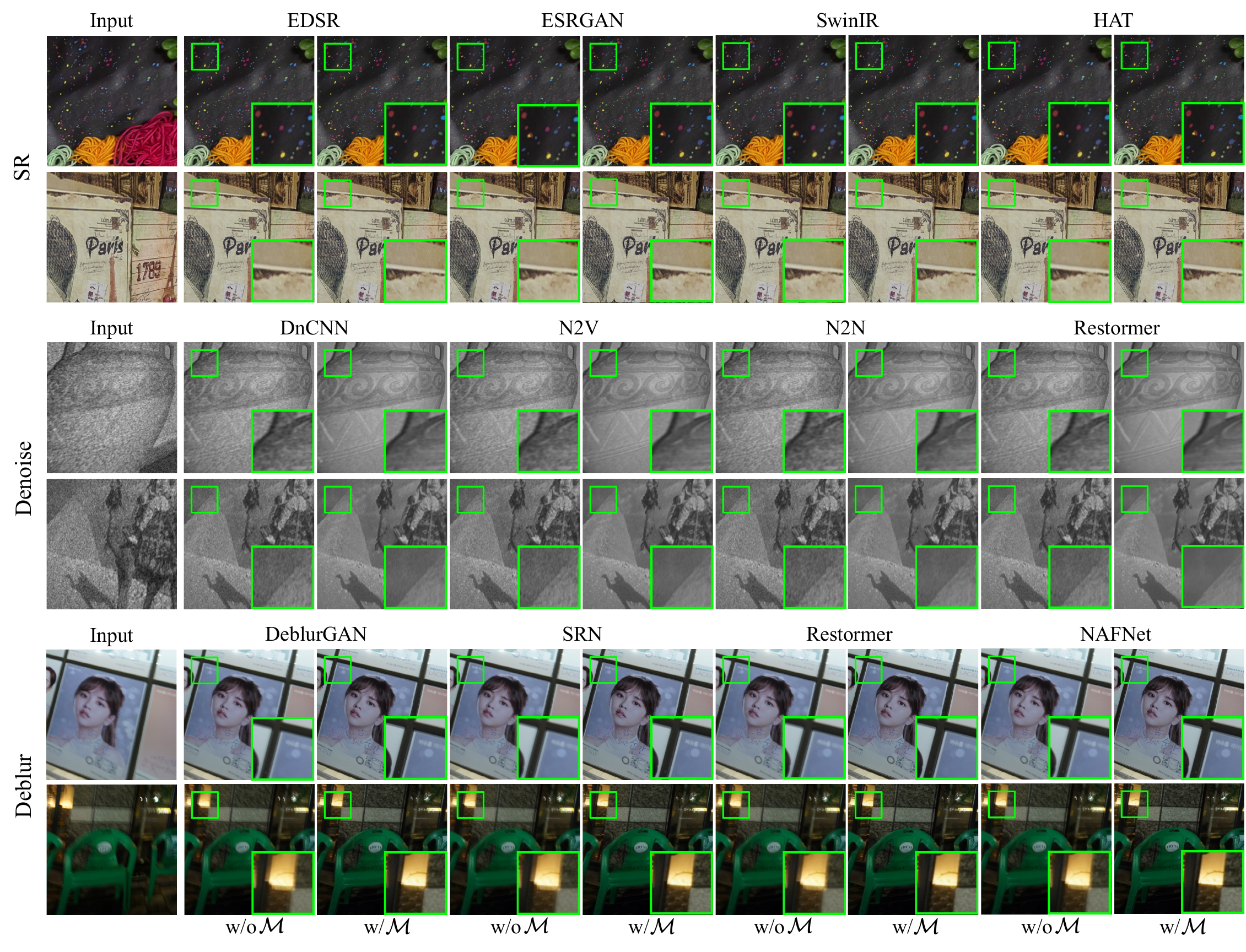}
    \caption{
    \textbf{Qualitative Restoration Results.} Restoration performance is compared with and without the matching network $\mathcal{M}_\theta$ across three tasks. For super-resolution (SR), we evaluate EDSR~\cite{Lim_2017_CVPR_Workshops}, ESRGAN~\cite{wang2018esrgan}, SwinIR~\cite{liang2021swinir}, and HAT~\cite{chen2023activating}; for denoising, we evaluate DnCNN~\cite{zhang2017beyond}, N2V~\cite{krull2019noise2void}, N2N~\cite{lehtinen2018noise2noiselearningimagerestoration}, and Restormer~\cite{Zamir2021Restormer}; and for deblurring, we evaluate DeblurGAN~\cite{DeblurGAN}, SRN~\cite{tao2018srndeblur}, Restormer~\cite{Zamir2021Restormer}, and NAFNet~\cite{chen2022simple}. The green box indicates the zoomed-in region.
    } 
    \label{fig:FIG_RESTORATION}
\end{figure*}

\begin{figure}[!htbp]
    \centering
    \includegraphics[width=1.0\columnwidth]{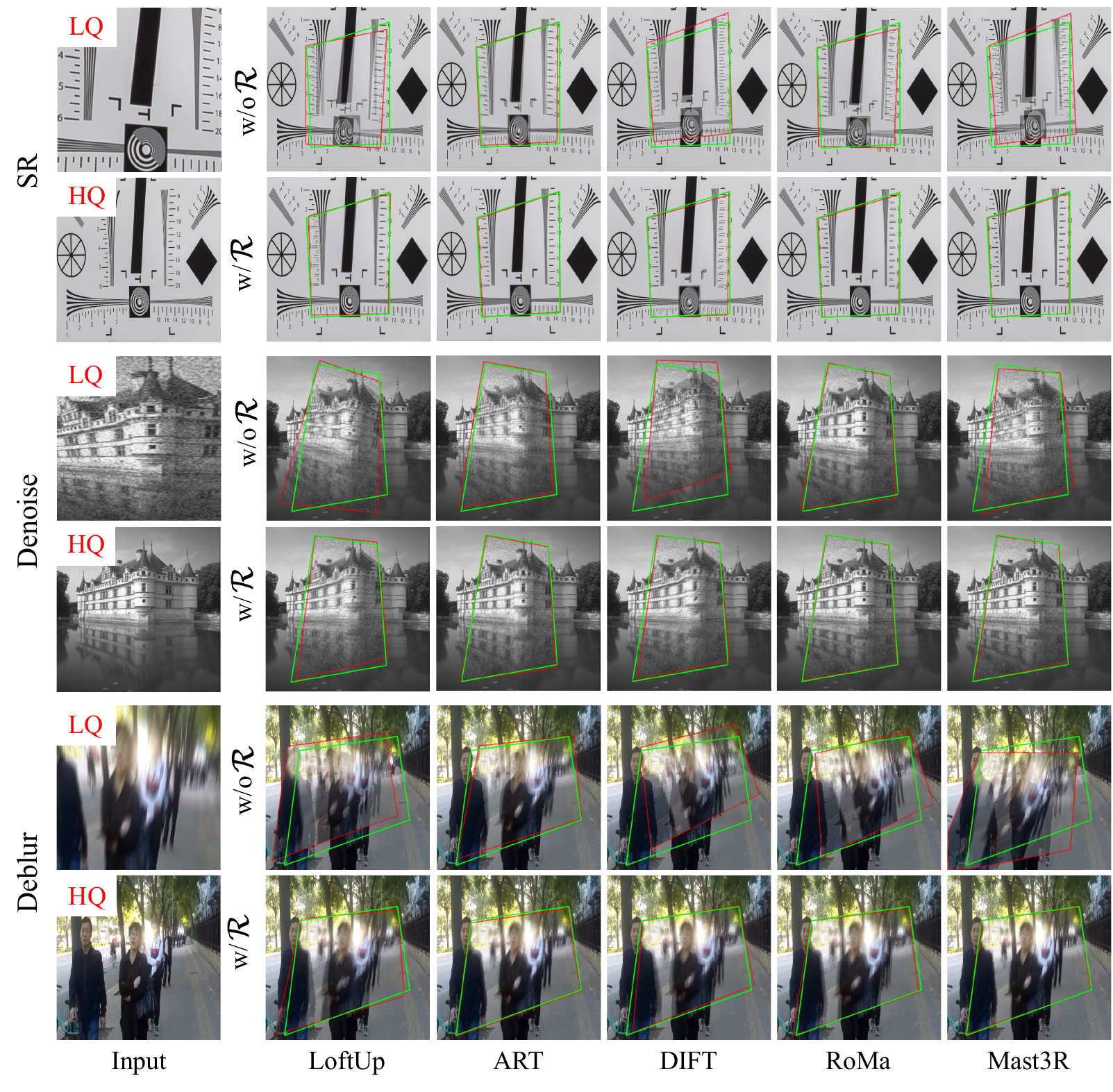}
    \caption{
    \textbf{Qualitative Matching Results.} Geometric transform estimation results for five matching networks (LoftUp~\cite{huang2025loftuplearningcoordinatebasedfeature}, ART~\cite{Lee_2025_ICCV}, DIFT~\cite{Tang2023EmergentCorrespondence}, RoMa~\cite{edstedt2024roma}, Mast3R~\cite{mast3r_eccv24}) across SR, denoising, and deblurring tasks, is evaluated with and without the restoration network $\mathcal{R}_\phi$. The green box denotes the ground-truth transform, while the red box represents the estimated transform.
    } 
    \label{fig:FIG_MATCHING}
\end{figure}

\begin{table*}[!htbp]
\centering
\caption{Experimental Results for Super-Resolution, Denoising, and Deblurring Tasks}
\label{tab:results}
\small
\renewcommand{\arraystretch}{0.9}
\resizebox{\textwidth}{!}{%
\begin{tabular}{lllcccccc}
\toprule
\multirow{2}{*}{\textbf{Task}} & \multirow{2}{*}{\textbf{Matcher} ($\mathcal{M}$)} & \multirow{2}{*}{\textbf{Restorator} ($\mathcal{R}$)} & \multicolumn{2}{c}{\textbf{Matching score}} & \multicolumn{4}{c}{\textbf{Restoration score}} \\
\cmidrule(lr){4-5} \cmidrule(lr){6-9}
 &  &  & \textbf{mAUC (w/o $\mathcal{R}$)} & \textbf{mAUC (w/ $\mathcal{R}$)} & \textbf{PSNR (w/o $\mathcal{M}$)} & \textbf{PSNR (w/ $\mathcal{M}$)} & \textbf{SSIM (w/o $\mathcal{M}$)} & \textbf{SSIM (w/ $\mathcal{M}$)} \\
\midrule
\multirow{20}{*}{SR} & \multirow{4}{*}{RoMa~\cite{edstedt2024roma}} & EDSR~\cite{Lim_2017_CVPR_Workshops} & $27.00$ & $36.60$ ($+9.61$) & $26.93$ & $31.72$ ($+4.80$) & $0.531$ & $0.747$ ($+0.216$) \\
 &  & ESRGAN~\cite{wang2018esrgan} & $24.46$ & $33.39$ ($+8.93$) & $26.40$ & $31.53$ ($+5.12$) & $0.505$ & $0.823$ ($+0.317$) \\
 &  & SwinIR~\cite{liang2021swinir} & $30.66$ & $34.36$ ($+3.70$) & $24.73$ & $28.28$ ($+3.55$) & $0.497$ & $0.688$ ($+0.191$) \\
 &  & HAT~\cite{chen2023activating} & $27.46$ & $31.79$ ($+4.33$) & $26.45$ & $29.87$ ($+3.42$) & $0.534$ & $0.716$ ($+0.183$) \\
\cline{2-9}
 & \multirow{4}{*}{ART~\cite{Lee_2025_ICCV}} & EDSR~\cite{Lim_2017_CVPR_Workshops} & $27.65$ & $35.93$ ($+8.28$) & $24.80$ & $29.34$ ($+4.54$) & $0.527$ & $0.670$ ($+0.143$) \\
 &  & ESRGAN~\cite{wang2018esrgan} & $28.86$ & $32.22$ ($+3.36$) & $24.26$ & $30.11$ ($+5.85$) & $0.552$ & $0.767$ ($+0.215$) \\
 &  & SwinIR~\cite{liang2021swinir} & $26.44$ & $29.22$ ($+2.78$) & $26.74$ & $31.06$ ($+4.32$) & $0.523$ & $0.762$ ($+0.239$) \\
 &  & HAT~\cite{chen2023activating} & $24.28$ & $33.55$ ($+9.27$) & $25.04$ & $30.02$ ($+4.99$) & $0.504$ & $0.736$ ($+0.232$) \\
\cline{2-9}
 & \multirow{4}{*}{DIFT~\cite{Tang2023EmergentCorrespondence}} & EDSR~\cite{Lim_2017_CVPR_Workshops} & $28.37$ & $31.85$ ($+3.48$) & $27.88$ & $33.20$ ($+5.33$) & $0.630$ & $0.862$ ($+0.232$) \\
 &  & ESRGAN~\cite{wang2018esrgan} & $28.78$ & $38.16$ ($+9.37$) & $24.35$ & $27.94$ ($+3.59$) & $0.462$ & $0.659$ ($+0.196$) \\
 &  & SwinIR~\cite{liang2021swinir} & $27.11$ & $31.28$ ($+4.17$) & $27.31$ & $31.39$ ($+4.07$) & $0.552$ & $0.776$ ($+0.224$) \\
 &  & HAT~\cite{chen2023activating} & $25.13$ & $33.54$ ($+8.42$) & $24.30$ & $30.26$ ($+5.96$) & $0.534$ & $0.710$ ($+0.177$) \\
\cline{2-9}
 & \multirow{4}{*}{LoftUp~\cite{huang2025loftuplearningcoordinatebasedfeature}} & EDSR~\cite{Lim_2017_CVPR_Workshops} & $24.04$ & $32.57$ ($+8.52$) & $26.83$ & $32.01$ ($+5.19$) & $0.590$ & $0.747$ ($+0.157$) \\
 &  & ESRGAN~\cite{wang2018esrgan} & $26.87$ & $29.79$ ($+2.93$) & $27.45$ & $32.32$ ($+4.87$) & $0.560$ & $0.754$ ($+0.194$) \\
 &  & SwinIR~\cite{liang2021swinir} & $26.49$ & $31.09$ ($+4.60$) & $26.92$ & $31.83$ ($+4.91$) & $0.604$ & $0.781$ ($+0.178$) \\
 &  & HAT~\cite{chen2023activating} & $24.96$ & $32.66$ ($+7.71$) & $27.04$ & $31.73$ ($+4.68$) & $0.595$ & $0.781$ ($+0.186$) \\
\cline{2-9}
 & \multirow{4}{*}{Mast3R~\cite{mast3r_eccv24}} & EDSR~\cite{Lim_2017_CVPR_Workshops} & $28.18$ & $33.60$ ($+5.42$) & $24.10$ & $27.43$ ($+3.32$) & $0.455$ & $0.675$ ($+0.220$) \\
 &  & ESRGAN~\cite{wang2018esrgan} & $26.51$ & $32.58$ ($+6.07$) & $27.63$ & $31.38$ ($+3.75$) & $0.572$ & $0.797$ ($+0.225$) \\
 &  & SwinIR~\cite{liang2021swinir} & $25.83$ & $28.45$ ($+2.62$) & $25.16$ & $28.64$ ($+3.48$) & $0.569$ & $0.726$ ($+0.158$) \\
 &  & HAT~\cite{chen2023activating} & $29.07$ & $38.04$ ($+8.97$) & $27.21$ & $30.77$ ($+3.56$) & $0.611$ & $0.759$ ($+0.148$) \\
\midrule
\multirow{25}{*}{Denoise} & \multirow{5}{*}{RoMa~\cite{edstedt2024roma}} & DnCNN~\cite{zhang2017beyond} & $62.92$ & $70.19$ ($+7.27$) & $28.59$ & $31.92$ ($+3.33$) & $0.597$ & $0.781$ ($+0.184$) \\
 &  & N2V~\cite{krull2019noise2void} & $63.09$ & $70.11$ ($+7.03$) & $27.03$ & $31.57$ ($+4.53$) & $0.592$ & $0.754$ ($+0.161$) \\
 &  & N2N~\cite{lehtinen2018noise2noiselearningimagerestoration} & $51.92$ & $55.28$ ($+3.36$) & $31.71$ & $35.68$ ($+3.97$) & $0.673$ & $0.884$ ($+0.211$) \\
 &  & Restormer~\cite{Zamir2021Restormer} & $55.82$ & $63.62$ ($+7.80$) & $31.81$ & $35.57$ ($+3.76$) & $0.672$ & $0.841$ ($+0.170$) \\
 &  & NAFNet~\cite{chen2022simple} & $54.56$ & $55.82$ ($+1.26$) & $30.05$ & $34.56$ ($+4.51$) & $0.601$ & $0.819$ ($+0.218$) \\
\cline{2-9}
 & \multirow{5}{*}{ART~\cite{Lee_2025_ICCV}} & DnCNN~\cite{zhang2017beyond} & $64.53$ & $67.21$ ($+2.68$) & $27.72$ & $32.19$ ($+4.47$) & $0.659$ & $0.768$ ($+0.109$) \\
 &  & N2V~\cite{krull2019noise2void} & $60.75$ & $67.09$ ($+6.33$) & $28.19$ & $33.37$ ($+5.18$) & $0.605$ & $0.831$ ($+0.226$) \\
 &  & N2N~\cite{lehtinen2018noise2noiselearningimagerestoration} & $60.14$ & $64.89$ ($+4.75$) & $27.45$ & $32.96$ ($+5.51$) & $0.589$ & $0.778$ ($+0.189$) \\
 &  & Restormer~\cite{Zamir2021Restormer} & $50.65$ & $55.79$ ($+5.14$) & $30.39$ & $33.44$ ($+3.05$) & $0.652$ & $0.791$ ($+0.139$) \\
 &  & NAFNet~\cite{chen2022simple} & $60.32$ & $62.54$ ($+2.22$) & $30.45$ & $34.61$ ($+4.16$) & $0.695$ & $0.806$ ($+0.111$) \\
\cline{2-9}
 & \multirow{5}{*}{DIFT~\cite{Tang2023EmergentCorrespondence}} & DnCNN~\cite{zhang2017beyond} & $55.46$ & $57.25$ ($+1.79$) & $31.62$ & $37.26$ ($+5.63$) & $0.645$ & $0.911$ ($+0.266$) \\
 &  & N2V~\cite{krull2019noise2void} & $63.08$ & $67.96$ ($+4.89$) & $29.65$ & $33.37$ ($+3.73$) & $0.599$ & $0.857$ ($+0.258$) \\
 &  & N2N~\cite{lehtinen2018noise2noiselearningimagerestoration} & $64.41$ & $69.84$ ($+5.43$) & $28.70$ & $32.74$ ($+4.05$) & $0.648$ & $0.845$ ($+0.197$) \\
 &  & Restormer~\cite{Zamir2021Restormer} & $64.19$ & $70.65$ ($+6.46$) & $30.21$ & $33.46$ ($+3.25$) & $0.614$ & $0.859$ ($+0.245$) \\
 &  & NAFNet~\cite{chen2022simple} & $59.70$ & $60.77$ ($+1.06$) & $27.51$ & $32.50$ ($+4.99$) & $0.558$ & $0.766$ ($+0.208$) \\
\cline{2-9}
 & \multirow{5}{*}{LoftUp~\cite{huang2025loftuplearningcoordinatebasedfeature}} & DnCNN~\cite{zhang2017beyond} & $58.78$ & $64.62$ ($+5.84$) & $30.26$ & $33.93$ ($+3.67$) & $0.670$ & $0.802$ ($+0.132$) \\
 &  & N2V~\cite{krull2019noise2void} & $55.21$ & $61.43$ ($+6.23$) & $30.25$ & $35.80$ ($+5.55$) & $0.664$ & $0.873$ ($+0.208$) \\
 &  & N2N~\cite{lehtinen2018noise2noiselearningimagerestoration} & $51.50$ & $55.07$ ($+3.57$) & $28.33$ & $32.06$ ($+3.73$) & $0.667$ & $0.780$ ($+0.113$) \\
 &  & Restormer~\cite{Zamir2021Restormer} & $64.27$ & $69.69$ ($+5.42$) & $30.97$ & $35.48$ ($+4.51$) & $0.667$ & $0.859$ ($+0.192$) \\
 &  & NAFNet~\cite{chen2022simple} & $53.12$ & $59.18$ ($+6.06$) & $28.40$ & $31.48$ ($+3.07$) & $0.636$ & $0.747$ ($+0.112$) \\
\cline{2-9}
 & \multirow{5}{*}{Mast3R~\cite{mast3r_eccv24}} & DnCNN~\cite{zhang2017beyond} & $65.05$ & $72.72$ ($+7.68$) & $31.57$ & $35.68$ ($+4.11$) & $0.620$ & $0.907$ ($+0.286$) \\
 &  & N2V~\cite{krull2019noise2void} & $56.85$ & $64.62$ ($+7.77$) & $31.82$ & $37.38$ ($+5.56$) & $0.652$ & $0.886$ ($+0.234$) \\
 &  & N2N~\cite{lehtinen2018noise2noiselearningimagerestoration} & $63.62$ & $66.84$ ($+3.22$) & $27.85$ & $32.52$ ($+4.67$) & $0.656$ & $0.820$ ($+0.164$) \\
 &  & Restormer~\cite{Zamir2021Restormer} & $59.12$ & $60.80$ ($+1.68$) & $30.08$ & $36.05$ ($+5.97$) & $0.610$ & $0.873$ ($+0.263$) \\
 &  & NAFNet~\cite{chen2022simple} & $64.04$ & $70.22$ ($+6.19$) & $30.49$ & $35.59$ ($+5.11$) & $0.638$ & $0.841$ ($+0.203$) \\
\midrule
\multirow{20}{*}{Deblur} & \multirow{4}{*}{RoMa~\cite{edstedt2024roma}} & DeblurGAN~\cite{DeblurGAN} & $21.33$ & $29.19$ ($+7.86$) & $26.07$ & $31.81$ ($+5.74$) & $0.523$ & $0.820$ ($+0.298$) \\
 &  & SRN~\cite{tao2018srndeblur} & $21.18$ & $28.08$ ($+6.90$) & $24.91$ & $30.30$ ($+5.39$) & $0.537$ & $0.766$ ($+0.229$) \\
 &  & Restormer~\cite{Zamir2021Restormer} & $15.26$ & $18.82$ ($+3.56$) & $24.05$ & $27.16$ ($+3.11$) & $0.478$ & $0.708$ ($+0.231$) \\
 &  & NAFNet~\cite{chen2022simple} & $14.01$ & $20.56$ ($+6.54$) & $20.21$ & $23.33$ ($+3.11$) & $0.437$ & $0.594$ ($+0.158$) \\
\cline{2-9}
 & \multirow{4}{*}{ART~\cite{Lee_2025_ICCV}} & DeblurGAN~\cite{DeblurGAN} & $11.78$ & $17.91$ ($+6.13$) & $25.39$ & $29.04$ ($+3.65$) & $0.520$ & $0.709$ ($+0.189$) \\
 &  & SRN~\cite{tao2018srndeblur} & $10.72$ & $16.91$ ($+6.19$) & $23.78$ & $28.70$ ($+4.91$) & $0.498$ & $0.790$ ($+0.292$) \\
 &  & Restormer~\cite{Zamir2021Restormer} & $17.23$ & $22.17$ ($+4.94$) & $25.57$ & $29.38$ ($+3.81$) & $0.505$ & $0.717$ ($+0.212$) \\
 &  & NAFNet~\cite{chen2022simple} & $10.35$ & $19.13$ ($+8.78$) & $25.85$ & $30.94$ ($+5.09$) & $0.508$ & $0.766$ ($+0.258$) \\
\cline{2-9}
 & \multirow{4}{*}{DIFT~\cite{Tang2023EmergentCorrespondence}} & DeblurGAN~\cite{DeblurGAN} & $12.19$ & $16.69$ ($+4.50$) & $23.84$ & $28.99$ ($+5.14$) & $0.493$ & $0.728$ ($+0.235$) \\
 &  & SRN~\cite{tao2018srndeblur} & $23.37$ & $30.80$ ($+7.43$) & $23.88$ & $28.72$ ($+4.84$) & $0.470$ & $0.718$ ($+0.248$) \\
 &  & Restormer~\cite{Zamir2021Restormer} & $14.98$ & $22.53$ ($+7.55$) & $20.10$ & $23.45$ ($+3.35$) & $0.357$ & $0.565$ ($+0.209$) \\
 &  & NAFNet~\cite{chen2022simple} & $21.98$ & $29.20$ ($+7.22$) & $23.32$ & $26.61$ ($+3.29$) & $0.466$ & $0.688$ ($+0.222$) \\
\cline{2-9}
 & \multirow{4}{*}{LoftUp~\cite{huang2025loftuplearningcoordinatebasedfeature}} & DeblurGAN~\cite{DeblurGAN} & $12.42$ & $18.03$ ($+5.60$) & $22.79$ & $27.64$ ($+4.85$) & $0.469$ & $0.670$ ($+0.201$) \\
 &  & SRN~\cite{tao2018srndeblur} & $15.24$ & $22.00$ ($+6.76$) & $23.52$ & $29.09$ ($+5.57$) & $0.486$ & $0.719$ ($+0.232$) \\
 &  & Restormer~\cite{Zamir2021Restormer} & $10.99$ & $17.84$ ($+6.85$) & $20.19$ & $24.94$ ($+4.76$) & $0.448$ & $0.656$ ($+0.208$) \\
 &  & NAFNet~\cite{chen2022simple} & $15.43$ & $22.29$ ($+6.86$) & $23.21$ & $27.84$ ($+4.64$) & $0.508$ & $0.710$ ($+0.201$) \\
\cline{2-9}
 & \multirow{4}{*}{Mast3R~\cite{mast3r_eccv24}} & DeblurGAN~\cite{DeblurGAN} & $23.46$ & $31.89$ ($+8.43$) & $21.37$ & $24.58$ ($+3.21$) & $0.387$ & $0.591$ ($+0.204$) \\
 &  & SRN~\cite{tao2018srndeblur} & $11.32$ & $18.42$ ($+7.10$) & $20.50$ & $24.46$ ($+3.96$) & $0.444$ & $0.589$ ($+0.144$) \\
 &  & Restormer~\cite{Zamir2021Restormer} & $21.40$ & $26.09$ ($+4.69$) & $20.83$ & $25.92$ ($+5.09$) & $0.429$ & $0.711$ ($+0.281$) \\
 &  & NAFNet~\cite{chen2022simple} & $20.29$ & $28.11$ ($+7.82$) & $21.97$ & $25.51$ ($+3.53$) & $0.465$ & $0.693$ ($+0.229$) \\
\bottomrule
\end{tabular}%
}
\end{table*}

\section{Experiments}
\subsection{Datasets}
We evaluate \textsc{MatRes} across three restoration tasks: super-resolution (SR), denoising, and deblurring, using widely adopted real-world and synthetic benchmarks.  
For SR, we use RealSR~\cite{cai2019toward}, DIV2K~\cite{Agustsson_2017_CVPR_Workshops}, Set5~\cite{Bevilacqua2012Set5}, Set14~\cite{Zeyde2010Set14}, Urban100~\cite{Huang2015Urban100}, and BSD100~\cite{Martin2001BSD}.  
Denoising is evaluated on Set5~\cite{Bevilacqua2012Set5}, Set14~\cite{Zeyde2010Set14}, and Urban100~\cite{Huang2015Urban100}.  
Deblurring is evaluated on GoPro~\cite{Nah_2017_CVPR}, HIDE~\cite{Shen_2019_ICCV}, RealBlur-J~\cite{rim_2020_ECCV}, RealBlur-R~\cite{rim_2020_ECCV}, and CelebA~\cite{liu2015faceattributes}.  

To simulate realistic conditions such as misalignment and lighting inconsistencies, we apply synthetic viewpoint shifts and illumination variations to all test sets.  
No training data are used, since \textsc{MatRes} performs test-time adaptation on pretrained models without any offline training.

\subsection{Baselines for Comparison}
We use several baselines for \textsc{MatRes} across SR, denoising, and deblurring as summarized in Tab.~\ref{tab:results}. 

For matching, we include diffusion-based method~\cite{Tang2023EmergentCorrespondence}, the auto-regressive method~\cite{Lee_2025_ICCV}, and transformer-based methods~\cite{edstedt2024roma,huang2025loftuplearningcoordinatebasedfeature,mast3r_eccv24}.

For restoration, we select representative task-specific networks to evaluate mutual guidance: EDSR~\cite{Lim_2017_CVPR_Workshops}, ESRGAN~\cite{wang2018esrgan}, 
SwinIR~\cite{liang2021swinir}, and HAT~\cite{chen2023activating} for SR; 
DnCNN~\cite{zhang2017beyond}, N2V~\cite{krull2019noise2void}, 
N2N~\cite{lehtinen2018noise2noiselearningimagerestoration}, 
Restormer~\cite{Zamir2021Restormer}, and NAFNet~\cite{chen2022simple} for denoising; 
and DeblurGAN~\cite{DeblurGAN}, SRN~\cite{tao2018srndeblur}, 
Restormer~\cite{Zamir2021Restormer}, and NAFNet~\cite{chen2022simple} for deblurring.

To benchmark against task-specific state-of-the-art methods, we additionally include 
AutoLUT~\cite{xu2025autolutlutbasedimagesuperresolution}, CATANet~\cite{liu2025CATANet}, 
and IM-LUT~\cite{park2025imlutinterpolationmixinglookup} for SR; 
MRPNet~\cite{Zamir2021MPRNet}, AFM~\cite{10657108}, and MAXIM~\cite{tu2022maxim} for denoising; 
Blur2Blur~\cite{pham2024blur2blur}, Stripformer~\cite{tsai2022stripformerstriptransformerfast}, 
and HI-DIFF~\cite{chen2023hierarchical} for deblurring; and MINIMA~\cite{ren2025minima}, 
DenseAffine~\cite{sun2025learning}, and MatchAnything~\cite{he2025matchanything} for matching.

\subsection{Implementation Details}\label{subsec:implementation}

\paragraph{Common setup}\label{subsec:setup}
We used the AdamW~\cite{loshchilov2019decoupled} optimizer with a learning rate of \(0.001\), \(\beta_1 = 0.9\), \(\beta_2 = 0.999\), and \(\epsilon = 10^{-8}\), applying weight decay every iterations with a decay rate of \(0.01\). 
The learning rate was halved every \(10\) iterations. 
The model $\mathcal{H}_\psi$ was trained for $\textsc{T}_{\text{adapt}}$ epochs using an NVIDIA A100 GPU, where $\textsc{T}_{\text{adapt}}$ was determined as the point at which the diffusion loss $\mathcal{L}_{\text{D}}$ no longer decreased.

\paragraph{Evaluation metric}
To evaluate alignment performance, we use the CEM approach~\cite{charles2003dual} to calculate the median error (MEE) and maximum error (MAE). 
The results are categorized as follows: 
i) \textit{Acceptable} (MAE \(<\) $50$ and MEE \(<\) $20$),
ii) \textit{Inaccurate} (others).
We calculated the mean value of Area Under Curve (AUC) score~\cite{hernandezmatas2017fire}.
In addition, to assess the performance of each enhancer, we measured the PSNR~\cite{5596999} and SSIM~\cite{5596999} between the EQ and HQ images.

\begin{table}[!htbp]
\centering
\footnotesize
\renewcommand{\arraystretch}{1.0}
\caption{Restoration score comparison across different tasks.}
\label{tab:unified_comparison}
\resizebox{\columnwidth}{!}{%
\begin{tabular}{llcccc}
\toprule
\textbf{Task} & \textbf{Dataset} & \multicolumn{4}{c}{\textbf{Restoration Score (PSNR / SSIM)}} \\
\midrule
\multicolumn{2}{l}{\textit{Methods:}} & \textbf{MatRes (ours)} & \textbf{AutoLUT}~\cite{xu2025autolutlutbasedimagesuperresolution} & \textbf{CATANet}~\cite{liu2025CATANet} & \textbf{IM-LUT}~\cite{park2025imlutinterpolationmixinglookup} \\
\cmidrule(lr){1-6}
\multirow{6}{*}{SR} & RealSR~\cite{cai2019toward} & $\mathbf{34.87}$ / $\mathbf{0.902}$ & $33.59$ / $0.864$ & $32.70$ / $0.839$ & $30.66$ / $0.777$ \\
 & DIV2K~\cite{Agustsson_2017_CVPR_Workshops} & $32.78$ / $0.758$ & $31.86$ / $0.733$ & $\mathbf{34.72}$ / $\mathbf{0.801}$ & $32.04$ / $0.700$ \\
 & Set5~\cite{Bevilacqua2012Set5} & $33.31$ / $0.742$ & $\mathbf{33.98}$ / $\mathbf{0.779}$ & $31.51$ / $0.700$ & $30.39$ / $0.700$ \\
 & Set14~\cite{Zeyde2010Set14} & $35.01$ / $0.941$ & $34.02$ / $0.917$ & $33.54$ / $0.899$ & $\mathbf{35.86}$ / $\mathbf{0.962}$ \\
 & Urban100~\cite{Huang2015Urban100} & $\mathbf{35.82}$ / $\mathbf{0.776}$ & $34.23$ / $0.758$ & $31.69$ / $0.700$ & $31.80$ / $0.732$ \\
 & BSD100~\cite{Martin2001BSD} & $33.80$ / $0.781$ & $33.14$ / $0.736$ & $31.48$ / $0.710$ & $\mathbf{34.85}$ / $\mathbf{0.795}$ \\
 & \textbf{Average} & $\mathbf{34.27}$ / $\mathbf{0.817}$ & $33.47$ / $0.798$ & $32.61$ / $0.775$ & $32.60$ / $0.778$ \\
\midrule
\multicolumn{2}{l}{\textit{Methods:}} & \textbf{MatRes (ours)} & \textbf{MRPNet}~\cite{Zamir2021MPRNet} & \textbf{AFM}~\cite{10657108} & \textbf{MAXIM}~\cite{tu2022maxim} \\
\cmidrule(lr){1-6}
\multirow{3}{*}{Denoise} & Set5~\cite{Bevilacqua2012Set5} & $\mathbf{38.26}$ / $\mathbf{0.891}$ & $36.46$ / $0.836$ & $37.49$ / $0.868$ & $36.98$ / $0.830$ \\
 & Set14~\cite{Zeyde2010Set14} & $\mathbf{38.54}$ / $\mathbf{0.850}$ & $37.85$ / $0.807$ & $35.72$ / $0.765$ & $35.86$ / $0.766$ \\
 & Urban100~\cite{Huang2015Urban100} & $36.64$ / $0.875$ & $\mathbf{37.41}$ / $0.889$ & $35.07$ / $0.790$ & $36.85$ / $\mathbf{0.905}$ \\
  & \textbf{Average} 
 & $\mathbf{37.81}$ / $\mathbf{0.872}$ 
 & $37.24$ / $0.844$
 & $36.09$ / $0.808$
 & $36.56$ / $0.834$ \\
\midrule
\multicolumn{2}{l}{\textit{Methods:}} & \textbf{MatRes (ours)} & \textbf{Blur2Blur}~\cite{pham2024blur2blur} & \textbf{Stripformer}~\cite{tsai2022stripformerstriptransformerfast} & \textbf{HI-DIFF}~\cite{chen2023hierarchical} \\
\cmidrule(lr){1-6}
\multirow{5}{*}{Deblur} & GoPro~\cite{Nah_2017_CVPR} & $30.91$ / $0.719$ & $\mathbf{31.69}$ / $\mathbf{0.750}$ & $29.50$ / $0.700$ & $31.56$ / $\mathbf{0.749}$ \\
 & HIDE~\cite{Shen_2019_ICCV} & $30.28$ / $0.733$ & $29.65$ / $0.700$ & $\mathbf{30.91}$ / $\mathbf{0.749}$ & $27.31$ / $0.700$ \\
 & RealBlur-J~\cite{rim_2020_ECCV} & $\mathbf{32.16}$ / $\mathbf{0.835}$ & $30.08$ / $0.764$ & $31.44$ / $0.748$ & $31.20$ / $0.812$ \\
 & RealBlur-R~\cite{rim_2020_ECCV} & $\mathbf{31.40}$ / $\mathbf{0.800}$ & $29.28$ / $0.766$ & $29.93$ / $0.705$ & $30.04$ / $0.771$ \\
 & CelebA~\cite{liu2015faceattributes} & $33.21$ / $0.871$ & $34.41$ / $\mathbf{0.903}$ & $32.48$ / $0.780$ & $\mathbf{34.55}$ / $0.895$ \\
  & \textbf{Average} 
 & $\mathbf{31.59}$ / $\mathbf{0.792}$ 
 & $31.02$ / $0.777$
 & $30.85$ / $0.736$
 & $30.93$ / $0.785$ \\
\bottomrule
\end{tabular}%
}
\end{table}

\begin{table}[!htbp]
\centering
\footnotesize
\renewcommand{\arraystretch}{1.0}
\caption{Matching score comparison across different tasks.}
\label{tab:matching_comparison}
\resizebox{\columnwidth}{!}{%
\begin{tabular}{llcccc}
\toprule
\multirow{2}{*}{\textbf{Task}} & \multirow{2}{*}{\textbf{Dataset}} & \multicolumn{4}{c}{\textbf{Matching Score (mAUC)}} \\
\cmidrule(lr){3-6}
& & \textbf{MatRes (ours)} & \textbf{MINIMA}~\cite{ren2025minima} & \textbf{DenseAffine}~\cite{sun2025learning} & \textbf{MatchAnything}~\cite{he2025matchanything} \\
\midrule
\multirow{7}{*}{SR} & RealSR~\cite{cai2019toward} & $\mathbf{34.85}$ & $24.99$ & $26.06$ & $29.60$ \\
 & DIV2K~\cite{Agustsson_2017_CVPR_Workshops} & $\mathbf{32.07}$ & $21.41$ & $26.38$ & $26.62$ \\
 & Set5~\cite{Bevilacqua2012Set5} & $\mathbf{31.38}$ & $23.18$ & $23.92$ & $25.05$ \\
 & Set14~\cite{Zeyde2010Set14} & $\mathbf{34.31}$ & $27.98$ & $27.38$ & $26.66$ \\
 & Urban100~\cite{Huang2015Urban100} & $\mathbf{35.88}$ & $27.77$ & $27.14$ & $31.51$ \\
 & BSD100~\cite{Martin2001BSD} & $\mathbf{34.37}$ & $29.85$ & $22.78$ & $22.65$ \\
  & \textbf{Average} 
 & $\mathbf{33.81}$ 
 & $25.86$
 & $25.94$
 & $27.01$ \\
\midrule
\multirow{4}{*}{Denoise} & Set5~\cite{Bevilacqua2012Set5} & $\mathbf{65.61}$ & $60.83$ & $56.14$ & $58.09$ \\
 & Set14~\cite{Zeyde2010Set14} & $\mathbf{57.95}$ & $53.67$ & $46.68$ & $51.88$ \\
 & Urban100~\cite{Huang2015Urban100} & $\mathbf{63.89}$ & $55.72$ & $55.51$ & $58.41$ \\
  & \textbf{Average} 
 & $\mathbf{62.48}$ 
 & $56.74$
 & $52.78$
 & $56.13$ \\
\midrule
\multirow{4}{*}{Deblur} & GoPro~\cite{Nah_2017_CVPR} & $\mathbf{28.33}$ & $22.76$ & $23.97$ & $21.73$ \\
 & HIDE~\cite{Shen_2019_ICCV} & $\mathbf{22.48}$ & $11.85$ & $15.62$ & $16.23$ \\
 & RealBlur-J~\cite{rim_2020_ECCV} & $\mathbf{23.94}$ & $13.52$ & $19.34$ & $12.04$ \\
 & RealBlur-R~\cite{rim_2020_ECCV} & $\mathbf{26.86}$ & $22.82$ & $16.34$ & $17.21$ \\
 & CelebA~\cite{liu2015faceattributes} & $\mathbf{24.85}$ & $10.05$ & $13.35$ & $15.91$ \\
  & \textbf{Average} 
 & $\mathbf{25.29}$ 
 & $16.60$
 & $17.72$
 & $16.62$ \\
\bottomrule
\end{tabular}%
}
\end{table}

\subsection{Evaluation of Mutual Guidance}
Tab.~\ref{tab:results} and the qualitative comparisons in Fig.~\ref{fig:FIG_RESTORATION} and Fig.~\ref{fig:FIG_MATCHING} collectively show that mutual guidance between the matching network $\mathcal{M}$ and the restoration network $\mathcal{R}$ consistently improves both restoration and matching across super-resolution, denoising, and deblurring. 

Fig.~\ref{fig:FIG_RESTORATION} demonstrates that incorporating $\mathcal{M}$ enables $\mathcal{R}$ to recover sharper structures with higher fidelity, and representative results in Tab.~\ref{tab:results} indicate PSNR improvements for SR methods~\cite{Lim_2017_CVPR_Workshops,wang2018esrgan}, with meaningful gains observed for denoising and deblurring models including Restormer~\cite{Zamir2021Restormer} and DeblurGAN~\cite{DeblurGAN}. 
These gains also lead to noticeable SSIM improvements, confirming that generative priors from $\mathcal{M}$ significantly enhance restoration. 

Fig.~\ref{fig:FIG_MATCHING} shows that cleaner features produced by $\mathcal{R}$ similarly improve geometric alignment accuracy. In the SR setting, representative mAUC improvements are observed for matchers such as RoMa~\cite{edstedt2024roma} and DIFT~\cite{Tang2023EmergentCorrespondence}, with comparably strong gains for Mast3R~\cite{mast3r_eccv24} in denoising and deblurring. 
The qualitative overlays further confirm that restored features lead to transform fields that align more closely with ground truth.

\subsection{Comparison with Task-Specific Baselines}
We additionally compare \textsc{MatRes} with task-specific state-of-the-art baselines as summarized in Tab.~\ref{tab:unified_comparison} and Tab.~\ref{tab:matching_comparison}. 
Because existing SR, denoising, and deblurring models cannot estimate geometric transforms, all baselines are evaluated using the transform predicted by \textsc{MatRes}. 

For restoration, \textsc{MatRes} achieves competitive or superior performance, including high scores on RealSR~\cite{cai2019toward} and Set14~\cite{Zeyde2010Set14}, outperforming strong SR methods such as AutoLUT~\cite{xu2025autolutlutbasedimagesuperresolution} and CATANet~\cite{liu2025CATANet}, denoising models such as MRPNet~\cite{Zamir2021MPRNet} and AFM~\cite{10657108}, and deblurring methods such as Blur2Blur~\cite{pham2024blur2blur} and Stripformer~\cite{tsai2022stripformerstriptransformerfast}. 

For matching, \textsc{MatRes} also achieves the highest mAUC across all datasets, clearly surpassing MINIMA~\cite{ren2025minima}, DenseAffine~\cite{sun2025learning}, and MatchAnything~\cite{he2025matchanything}, with notable improvements for challenging datasets. 
These results confirm that \textsc{MatRes} provides strong improvements to both restoration and geometric alignment compared with task-specific state-of-the-art methods.

\section{Conclusion}
We introduced \textsc{MatRes}, a practical and offline-training-free framework for jointly improving image restoration and geometric matching from image pairs with different qualities and viewpoints. By leveraging a realistic setting in which one image serves as a high-quality reference, \textsc{MatRes} enforces conditional similarity between observations and enables pretrained restoration and matching networks to guide each other during test-time adaptation. Experiments show that \textsc{MatRes} consistently enhances both restoration quality and matching accuracy compared with using either component alone, suggesting a promising direction for unified test-time adaptation and joint multi-view enhancement across diverse image domains.

{
    \small
    \bibliographystyle{unsrt}
    \bibliography{main}
}


\end{document}